\documentclass[10pt,twocolumn,letterpaper]{article}

\usepackage{cvpr}
\usepackage{epsfig}
\usepackage{amsmath}
\usepackage{graphicx,amssymb,icomma}
\usepackage{subfigure}
\usepackage{times}
\usepackage{latexsym}
\usepackage{algorithm2e}
\usepackage{amsmath}
\usepackage{url}
\usepackage{placeins}
\usepackage{caption}
\usepackage{footnote}
\usepackage{booktabs}
\usepackage{lipsum}
\usepackage{appendix}
\usepackage{colortbl}
\usepackage{bm}
\usepackage{multirow}
\usepackage{algorithmicx}
\usepackage{listings}
\usepackage[noend]{algpseudocode}
\usepackage{abstract}
\usepackage{enumitem}
\usepackage{chngcntr}
\usepackage{etoolbox}
\usepackage{tablefootnote}
\usepackage[symbol]{footmisc}

\usepackage[pagebackref=true,breaklinks=true,letterpaper=true,colorlinks,bookmarks=false]{hyperref}

\newcommand{\specialcell}[2][c]{\begin{tabular}[#1]{@{}c@{}}#2\end{tabular}}

\newcommand{\ft}[1]{\fontsize{#1pt}{1em}\selectfont}

\cvprfinalcopy 

\def\cvprPaperID{****} 

\ifcvprfinal\fi
\begin{document}

\title{Learning Sparse Mixture of Experts for Visual Question Answering}


\author{Vardaan Pahuja$^{\dag\S}$\thanks{Corresponding author: Vardaan Pahuja $<$vardaanpahuja@gmail.com$>$} \qquad Jie Fu$^{\dag\ddag}$ \qquad Christopher J. Pal$^{\dag\ddag}$\\~\\
$^{\dag}$Mila \qquad $^{\S}$Universit\'e de Montr\'eal \qquad $^\ddag$Polytechnique Montr\'eal}

\maketitle

\begin{abstract}
    There has been a rapid progress in the task of Visual Question Answering with improved model architectures. Unfortunately, these models are usually computationally intensive due to their sheer size which poses a serious challenge for deployment. We aim to tackle this issue for the specific task of Visual Question Answering (VQA). A Convolutional Neural Network (CNN) is an integral part of the visual processing pipeline of a VQA model (assuming the CNN is trained along with entire VQA model). In this project, we propose an efficient and modular neural architecture for the VQA task with focus on the CNN module. Our experiments demonstrate that a sparsely activated CNN based VQA model achieves comparable performance to a standard CNN based VQA model architecture.
\end{abstract}



\section{Introduction}
Our main goal is to optimize the convolutions performed by the Convolutional Neural Network (CNN) in a conditional computation setting. We use an off-the-shelf neural architecture for both VQA v2 \cite{balanced_vqa_v2} and CLEVR \cite{johnson2017clevr} datasets and just replace the CNN with a modularized version of the ResNeXt \cite{xie2017aggregated} CNN as we describe below. The details of convolutional architecture used in the VQA v2 model and CLEVR model are illustrated in Table~\ref{tab:cnn_vqa} and Table~\ref{tab:cnn_clevr} respectively in Appendix~\ref{sec:cnn_layout}.

\section{Bottleneck convolutional block}
\begin{figure}[!ht]
\includegraphics[width=\columnwidth]{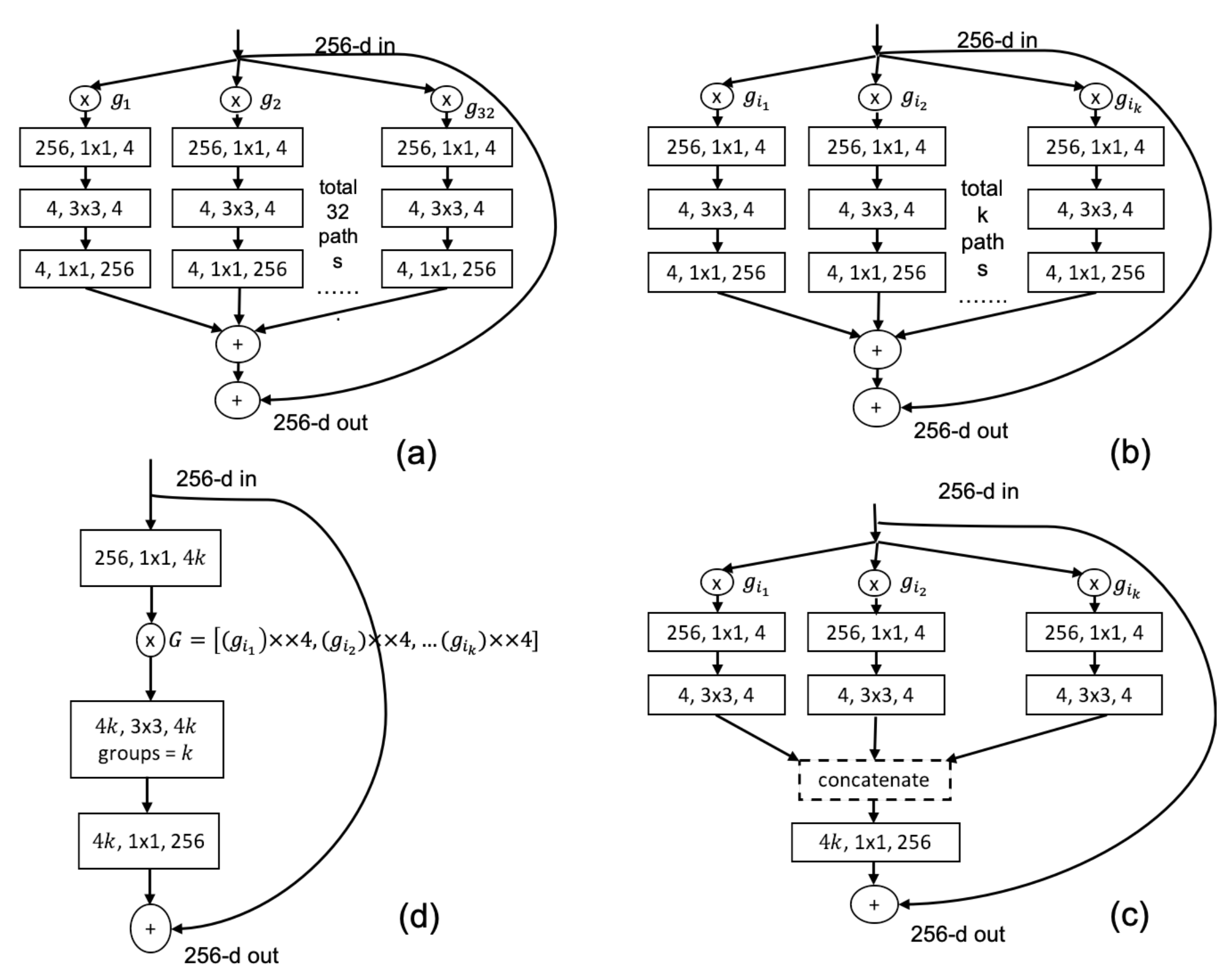}
\caption{Architecture of residual block of conditional gated ResNeXt-101 ($32\times 4d$) assuming we choose to turn ON paths with $top-k$ gating weights. Here $i_1,\cdots,i_k$ denote the indices of groups in $top-k$.}
\label{resnext_w}
\end{figure}

Figure~\ref{resnext_w} (a-d) shows the transformation of modularized ResNeXt-101 $(32\times4d)$ residual block to its grouped convolution form. This technique is similarly applicable for the convolutional block used for CLEVR dataset. We introduce a gating mechanism to assign weights to each of the paths (which equal 32 in the example shown). We treat each path as a convolutional module which should potentially be used for a specific function. The gate values are normalized to sum to unity and are conditioned on the LSTM based feature representation of the question. The working of gate controller is detailed in section~\ref{sec:gate_c}. See Figure~\ref{resnext_orig} (Appendix~\ref{sec:vqa_arch}) for the original ResNeXt-101 $(32\times4d)$ residual block.

In order to optimize the computation of a ResNeXt residual block, we execute just the top-$k$ (out of 32) paths and zero out the contribution of others. This is based on the hypothesis that the gate controller shall determine the most important modules (\textit{aka} paths) to execute by assigning higher weights to more important modules. 
In our efficient implementation, we avoid executing the groups which don't fall in top-$k$. More technically, we aggregate the non-contiguous groups of the input feature map, which fall in top-$k$, into a new feature map. We perform the same trick for the corresponding convolutional and batch-norm weights and biases.
\newpage

Computational complexity of the ResNeXt convolutional block (in terms of floating point operations)\footnote{No. of FLOPS of a convolutional block (no grouping) = $C_{in}*C_{out}*p*p*H_o*W_o$, No. of FLOPS of a convolutional block (with grouped convolution) = $\frac{C_{in}*C_{out}*p*p*H_o*W_o}{k}$ where $C_{in}$, $C_{out}$, $p$, $k$, $H_o$, $W_o$ denote the number of input channels, no. of output channels, kernel size, no. of groups, output feature map height and width respectively}=
\vspace{-1ex}
\begin{equation*}
\begin{split}
&= \textrm{complexity(\textit{conv-reduce})\footnotemark}+ \textrm{complexity(\textit{conv-conv})}\\
&+\textrm{complexity(\textit{conv-expand})}\\ &+\textrm{complexity(\textit{bn-reduce, bn, bn-expand})}\\
&=C * (k*d) * 1 * 1 * H^1_o * W^1_o  +\\  &\frac{(k*d) * (k*d) * 3 * 3}{k} * H^2_o * W^2_o
\\&+ (k*d) * C * 1 * 1 * H^3_o * W^3_o + O(k)\\
&= k * d * C * H^1_o * W^1_o + k * d * C * H^3_o * W^3_o\\
&+ 9 * d^2 * k * H^2_o * W^2_o + O(k)
\end{split}
\end{equation*}
\footnotetext{Here, $d$ denotes the size of group in group convolution (equals 4 for Figure~\ref{resnext_w}) and $C$
denotes the no. of channels in the feature map input to the convolutional block.}
\textbf{Notation}: \textit{conv-reduce}, \textit{conv-conv} and \textit{conv-expand} denote the $1\times1$, $3\times3$ and $1\times1$ convolutional layers in a ResNeXt convolutional block (in that order).

The implementation of modularized ResNeXt block is more efficient than the regular implementation in the case when $k < 32$.
The comparison of FLOPS with varying values of the hyper-parameter $k$ is shown in Table~\ref{results_vqa} for the VQA v2 model and 
Table~\ref{results_clevr} for the CLEVR model.

\subsection{Working of gate controller}\label{sec:gate_c}
The function of the gate controller is to choose the set of experts which are the most important. The gate controller predicts a soft attention \cite{yang2016stacked} over the image grid by combining visual and question features. The weighted visual feature vector is then summed with the question feature representation to get a query vector in the multimodal space. This new query vector is fed to an MLP which predicts attention weights for the set of experts. The experts whose weights are contained in top-$k$ are selected for execution and their outputs are weighted by the gate values assigned. See Appendix~\ref{sec:gate_controller} for more details on the attention mechanism.

\section{Experiments}
\subsection{VQA v2 dataset}
We use the Bottom-up attention model for VQA v2 dataset as proposed in \cite{teney2017tips} as our base model and replace the CNN sub-network with our custom CNN. A schematic diagram to illustrate the working of this model is given in Figure~\ref{arch_vqa} (Appendix~\ref{sec:vqa_arch}). The results (see Table~\ref{results_vqa}) show that there is a very minimal loss in accuracy from  $0 \%$ sparsity to $50 \%$ sparsity. However, with $75 \%$ sparsity\footnote{Here, $75 \%$ sparsity means that $75 \%$ of the modules/paths in the ResNeXt convolutional block are turned off.}, there is a marked $3.62 \%$ loss in overall accuracy.

\subsection{CLEVR dataset}
We use the Relational Networks model \cite{santoro2017simple} because it is one of the few models which is fully-supervised and trains the CNN in the main model pipeline. We replace the Vanilla CNN used in their model with our modularized CNN and report the results on the CLEVR dataset. A diagram to illustrate the working of this model is shown in Figure~\ref{rel_net} (Appendix~\ref{sec:vqa_arch}).  The CNN used for this model has four layers with one residual ResNeXt block each followed by a $1\times1$ convolutional layer. The results (see Table~\ref{results_clevr}) show that with a slight dip in performance, the model which uses $50 \%$ sparsity has comparable performance with the one which doesn't have sparsity in the convolutional ResNeXt block. 

\begin{table}[!ht]
\centering
\resizebox{0.99\linewidth}{!}{
\begin{tabular}{c|c|c}
\toprule
\textbf{Architecture for CNN}                 & \specialcell{\textbf{FLOPS}\\\textbf{(CNN)}}\tablefootnote{The FLOPS calculation assumes an input image of size $224\times224\times3$} & \textbf{Acc. (\%)} \\ \hline
\specialcell{ResNeXt-32 (101 x 32d)\\(baseline)}\tablefootnote{The baseline model doesn't use R-CNN based features, so the accuracy is not directly comparable with state of the art approaches.}                   & 156.04E+09 & 54.51                  \\
\specialcell{Modular ResNeXt-32 (101 x 32d)\\k = 32 (0 $\%$ sparsity)} & 181.39E+09 & 54.90                  \\
\specialcell{Modular ResNeXt-32 (101 x 32d)\\k = 16 (50 $\%$ sparsity)} & 77.72E+09 & 54.47                  \\
\specialcell{Modular ResNeXt-32 (101 x 32d)\\k = 8 (75 $\%$ sparsity)} & 45.94E+09 & 51.28                  \\ \bottomrule
\end{tabular}}
\caption{Results on VQA v2 validation set}
\label{results_vqa}
\vspace{-2ex}
\end{table}

\vspace{-2ex}
\begin{table}[!ht]
\centering
\resizebox{0.99\linewidth}{!}{
\begin{tabular}{c|c|c}
\toprule
\textbf{CNN Model description}            & \textbf{FLOPS (CNN)\tablefootnote{The FLOPS calculation assumes an input image of size $128\times128\times3$}} & \textbf{Val. Acc. (\%)} \\ \hline
\specialcell{Modular CNN, k=12\\(baseline)}     & 5.37E+07 & 94.05              \\
\specialcell{Modular CNN, k=6,\\50 \% sparsity} & 3.21E+07 & 92.23              \\ \bottomrule
\end{tabular}}
\caption{Results on CLEVR v1.0 validation set (Overall accuracy)}
\label{results_clevr}
\end{table}

\section{Conclusion}
We presented a general framework for utilizing conditional computation to sparsely execute a subset of modules in a convolutional block of ResNeXt model. The amount of sparsity is a user-controlled hyper-parameter which can be used to turn off the less important modules conditioned on the question representation, thereby increasing the computational efficiency. Future work may include studying the utility of this technique in other multimodal machine learning applications which support use of conditional computation.

{\small
\bibliographystyle{ieee_fullname}
\bibliography{egbib}
}
\appendix
\section{Related Work}
\textbf{Mixture of Experts} Mixture of Experts \cite{jacobs1991adaptive} is a formulation in machine learning which employs the divide-and-conquer principle to solve a complex problem by dividing the neural network into several expert networks. In the Mixture of MLP Experts (MME) method \cite{waterhouse1998classification, nguyen2006cooperative, ebrahimpour2008view}, a gating network is used to assign weights to the outputs of the corresponding expert networks. The gating network is an MLP followed by softmax operator. The final output of the ensemble is a weighted sum of outputs of each of the individual expert networks.
\cite{jo2018modularity} uses a mixture of experts for visual reasoning tasks in which each expert is a stack of residual blocks.

\textbf{Conditional Computation} Conditional Computation is the technique of activating only a sub-portion of the neural network depending on the inputs. For instance, if a visual question answering system has to count the number of a specified object \textit{vs} if it has to tell the color of an object, the specific features needed to give the correct answer in either case are different. Hence, there is a potential to reduce the amount of computation the network has to perform in each case, which can be especially useful to train large deep networks efficiently. The use of stochastic neurons with binary outputs to selectively turn-off experts in a neural network has been explored in \cite{bengio2013estimating}. \cite{davis2013low} uses a low-rank approximation of weight matrix of MLP to compute the sign of pre-nonlinearity activations. In case of ReLU activation function, it is then used to optimize the matrix multiplication for the MLP layer. \cite{bengio2015conditional} uses Policy gradient to sparsely activate units in feed-forward neural networks by relying on conditional computation. \cite{shazeer2017outrageously} proposes Sparsely-Gated Mixture-of-Experts layer (MoE) which uses conditional computation to train huge capacity models on low computational budget for language modeling and machine translation tasks. It makes use of noisy top-$K$ mechanism in which a random noise is added to gating weights and then top-$K$ weights are selected. Another line of work makes use of conditional computation in VQA setting. DPPNet \cite{noh2015dppnet} makes use of a dynamic parameter layer (fully-connected) conditioned on the question representation for VQA. ABC-CNN \cite{Chen2015ABCCNNAA} predicts convolutional kernel weights using an MLP which takes question representation as input. Here, the advantage is that the question conditioned convolutional kernels can filter out unrelated image regions in the visual processing pipeline itself.

\textbf{Computationally efficient CNNs} \cite{huang2017multi} proposes Multi-Scale Dense Convolutional Networks (MSDNet) to address two key issues in (i) budgeted classification- distribute the computation budget unevenly across easy and hard examples, (ii) anytime prediction: the network can output the prediction result at at any layer depending on the computation budget without significant loss in accuracy. The optimization of computational complexity of CNNs at inference-time has been studied in \cite{bolukbasi2017adaptive} in which an adaptive early-exit strategy is learned to by-pass some of the network's layers in order to save computations. MobileNet \cite{howard2017mobilenets} uses depth-wise separable convolutions to build light weight CNNs for deployment on mobile devices. 

\textbf{CNN architectures} The invention of Convolutional Neural Networks (CNNs) has led to a remarkable improvement in performance for many computer vision tasks \cite{krizhevsky2012imagenet, long2015fully, ren2015faster}. In recent years, there has been a spate of different CNN architectures with changes to depth \cite{simonyan2014very}, topology \cite{he2016deep, szegedy2015going}, etc. The use of \textit{split-transform-merge} strategy for designing convolutional blocks (which can be stacked to form the complete network) has shown promise for achieving top performance with a lesser computational complexity \cite{szegedy2015going, szegedy2017inception, szegedy2016rethinking, xie2017aggregated}. The ResNeXt \cite{xie2017aggregated} CNN model proposes cardinality (size of set of transformations in a convolutional block) as another dimension apart from depth and width to investigate, for improving the performance of convolutional neural networks. Squeeze-and-Excitation Networks \cite{hu2017squeeze} proposes channel-wise attention in a convolutional block and helped improve the state of art in ILSVRC 2017 classification competition.

\textbf{Grouped Convolution} In grouped convolution, each filter convolves only with the input feature maps in its group. The use of grouped convolutions was first done in AlexNet \cite{krizhevsky2012imagenet} for training a large network on 2 GPUs. A recently proposed CNN architecture named CondenseNet \cite{huang2018condensenet} makes use of learned grouped convolutions to minimise superfluous feature-reuse and achieve computational efficiency.

\section{Gate controller}\label{sec:gate_controller}
The gate controller takes as input the LSTM based representation of the question and the intermediate convolutional map which is the output of the previous block. Given the image features $\bm{v_I}$ and question features $\bm{v_Q}$, we perform the fusion of these features, followed by a linear layer and softmax to generate the attention $\bm{p^I}$ over the pixels of the image feature input.

\begin{equation*}
\begin{split}
\bm{h_A} &= tanh(\bm{W_{I,A}}\bm{v_I} \oplus (\bm{W_{Q,A}}\bm{v_Q} + \bm{b_A})) \\
\bm{p^I} &= softmax(\bm{W_P} \bm{h_A} + \bm{b_P}) \\
\end{split}
\end{equation*}
This pixel-wise attention is then used to modulate the image features and the resulting feature vector is summed with the question feature vector to obtain the combined query vector in multi-modal space. The gating weights are obtained by an MLP followed by ReLU (Recified Linear Unit) activation on the query vector and subsequent L1 normalization.

\begin{equation*}
\begin{split}
\bm{\tilde{v}_I} &= \sum_i p^I_i \bm{v_i} \\
\bm{u_{query}} &= \bm{\tilde{v}_I} + \bm{v_Q} \\
\bm{g} &= \bm{W_g} \bm{u_{query}} + \bm{b_g} \\
\bm{g^{'}} &= \frac{ReLU(\bm{g})}{||ReLU(\bm{g})||_1} \\
\end{split}
\end{equation*}
Notation: $ \bm{W_I} \in \mathbb{R}^{A \times B}$,
$ \bm{W_{I,A}} \in \mathbb{R}^{C \times B}$,
$ \bm{W_{Q,A}} \in \mathbb{R}^{B \times C}$,
$ \bm{W_P} \in \mathbb{R}^{1 \times C}$,
$ \bm{b_P} \in \mathbb{R}^{1 \times D}$,
$ \bm{W_g} \in \mathbb{R}^{B \times E}$,
$ \bm{b_g} \in \mathbb{R}^{B \times 1}$,
$ A = \text{Feature dim. of each img. region in final conv. layer} $,
$ B = \text{question feature len(from LSTM/GRU)} $,
$ C = \text{Hidden layer dim.} $,
$ D = \text{No. of img. regions} $,
$ E = \text{No. of modules/residual blocks} $

\section{Additional Training details}
We add an additional loss term which equals the square of coefficient of variation (CV) of gate values for each convolutional block. 
$$ CV(\bm{g}) = \frac{\sigma(\bm{g})}{\mu(\bm{g})}$$
This helps to balance out the variation in gate values \cite{shazeer2017outrageously} otherwise the weights corresponding to the modules which get activated initially will increase in magnitude and this behavior reinforces itself as the training progresses.

\section{CNN layouts}\label{sec:cnn_layout}
\newcommand{\blockxvqa}[3]{\multirow{3}{*}{
\(\left[
\begin{array}{l}
\text{1$\times$1, #2}\\
[-.1em] \text{3$\times$3, #2, $C$=32}\\
[-.1em] \text{1$\times$1, #1}\\
\end{array}\right]\)$\times$#3}
}

\renewcommand\arraystretch{1.25}
\setlength{\tabcolsep}{1.2pt}
\begin{table}[t]
\begin{center}
\footnotesize
\begin{tabular}{c|c|c}
\hline
 stage & output & Description \\
\hline
conv1 & \ft{7} 112$\times$112 & 7$\times$7, 64, stride 2 \\
\hline
\multirow{4}{*}{conv2} & \multirow{4}{*}{\ft{7} 56$\times$56} & 3$\times$3 max pool, stride 2 \\\cline{3-3}
  &  &  \blockxvqa{256}{128}{3}\\
  &  & \\
  &  & \\
\hline
\multirow{3}{*}{conv3} &  \multirow{3}{*}{\ft{7} 28$\times$28} 
  & \blockxvqa{512}{256}{4}\\
  &  & \\
  &  & \\
\hline
\multirow{3}{*}{conv4} & \multirow{3}{*}{\ft{7} 14$\times$14} 
  & \blockxvqa{1024}{512}{23}\\
  &  & \\
  &  & \\
\hline
\multirow{3}{*}{conv5} & \multirow{3}{*}{\ft{7} 7$\times$7} 
& \blockxvqa{2048}{1024}{3}\\
  &  & \\
  &  & \\
\hline
\end{tabular}
\end{center}
\caption{Modular CNN for VQA v2 model}
\label{tab:cnn_vqa}
\vspace{-.5em}
\end{table}
\newcommand{\blockx}[3]{\multirow{3}{*}{
\(\left[
\begin{array}{l}
\text{1$\times$1, #2}\\
[-.1em] \text{3$\times$3, #2, $C$=12}\\
[-.1em] \text{1$\times$1, #1}\\
\end{array}\right]\)$\times$#3}
}

\renewcommand\arraystretch{1.25}
\setlength{\tabcolsep}{1.2pt}
\begin{table}[t]
\begin{center}
\footnotesize
\begin{tabular}{c|c|c}
\hline
stage & output & Description \\
\hline
conv1 & \ft{7} 64$\times$64 & 3$\times$3, 64, stride 2 \\
\hline
\multirow{4}{*}{conv2} & \multirow{4}{*}{\ft{7} 32$\times$32} & 3$\times$3 max pool, stride 2 \\\cline{3-3}
  &  & \blockx{48}{48}{1}\\
  &  & \\
  &  & \\
\hline
\multirow{3}{*}{conv3} &  \multirow{3}{*}{\ft{7} 16$\times$16} 
  &  \blockx{48}{48}{1}\\
  &  & \\
  &  & \\
\hline
\multirow{3}{*}{conv4} & \multirow{3}{*}{\ft{7} 8$\times$8} 
  &  \blockx{48}{48}{1}\\
  &  & \\
  &  & \\
\hline
conv5 & {\ft{7} 8$\times$8} & \specialcell{$1\times1$ conv. layer\\with 24 o/p channels}\\
\hline
\end{tabular}
\end{center}
\caption{Modular CNN for Relational Networks Model}
\label{tab:cnn_clevr}
\vspace{-.5em}
\end{table}
\FloatBarrier
\section{VQA Model architectures}\label{sec:vqa_arch}

\begin{figure*}[!ht]
\includegraphics[width=2\columnwidth]{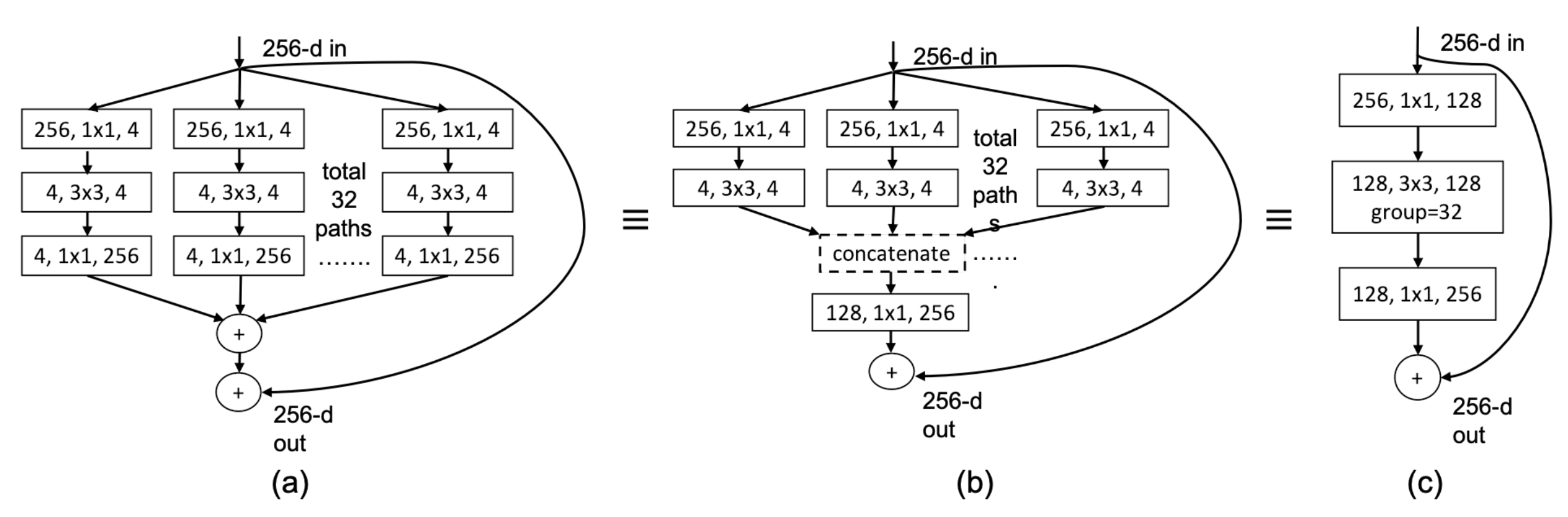}
\caption{Architecture of a sample block of ResNeXt-101 ($32\times 4d$)}
\label{resnext_orig}
\end{figure*}

\begin{figure*}[!ht]
\includegraphics[width=2\columnwidth]{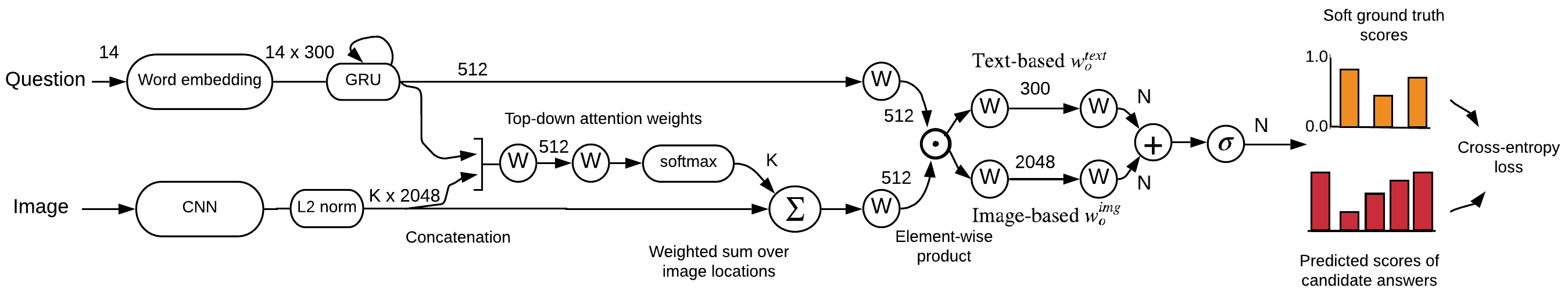}
\caption{Model architecture for VQA v2 dataset (adapted from \cite{teney2017tips})}
\label{arch_vqa}
\end{figure*}

\begin{figure*}[!ht]
\includegraphics[width=2\columnwidth]{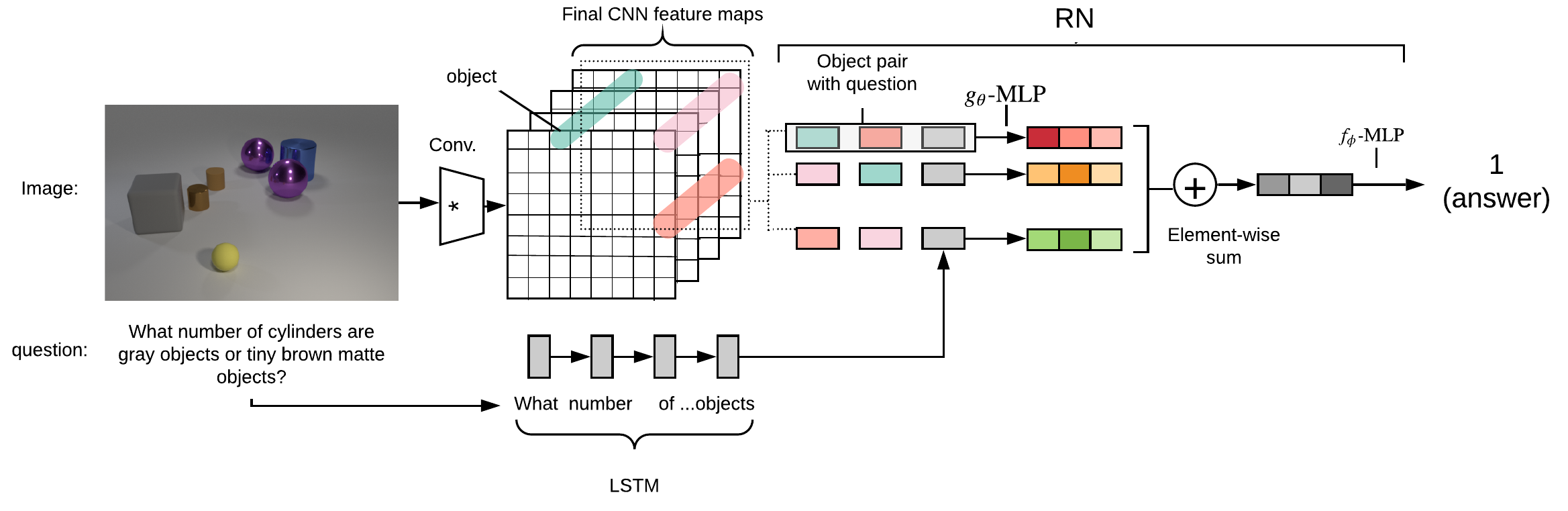}
\caption{Model architecture for Relational Networks (adapted from \cite{santoro2017simple})}
\label{rel_net}
\end{figure*}

\end{document}